# Constructing the F-Graph with a Symmetric Constraint for Subspace Clustering

Kai Xu, Xiao-Jun Wu*, Wen-Bo Hu

School of IoT Engineering, Jiangnan University, Wuxi 214122, China

**Abstract**: Based on further studying the low-rank subspace clustering (LRSC) and L2-graph subspace clustering algorithms, we propose a F-graph subspace clustering algorithm with a symmetric constraint (FSSC), which constructs a new objective function with a symmetric constraint basing on F-norm, whose the most significant advantage is to obtain a closed-form solution of the coefficient matrix. Then, take the absolute value of each element of the coefficient matrix, and retain the k largest coefficients per column, set the other elements to 0, to get a new coefficient matrix. Finally, FSSC performs spectral clustering over the new coefficient matrix. The experimental results on face clustering and motion segmentation show FSSC algorithm can not only obviously reduce the running time, but also achieve higher accuracy compared with the state-of-the-art representation-based subspace clustering algorithms, which verifies that the FSSC algorithm is efficacious and feasible.

## 1. INTRODUCTION

High dimensional datasets become increasingly ubiquitous with the rapid development of information technology over the past few decades, which not only increases the demand for memory and algorithm running time but also have an adverse effect on the performance of algorithms [1-4] due to the "curse of dimensionality". To reduce the dimensionality of original images, many subspace learning are presented [5-7].

In many application fields, high-dimensional data in the same class or directory can be well represented by low-dimensional subspaces. In recent years, subspace clustering base on sparse representation has received widespread attention due to its wide application in signal processing [8-9].

Subspace clustering algorithms are based on the idea that the intrinsic dimension of high dimensional datasets is often much smaller than the dimension of the ambient space, whose task is to segment all data points into their respective subspaces, which have lots of applications in computer vision, image processing, and systems theory. The existing subspace clustering algorithms can be divided into four main categories: iterative methods [10, 11], algebraic methods [12, 13], statistical methods [14, 15], and spectral clustering-based methods [16-18]. The performance of spectral clustering-based methods is excellent for some applications in computer vision [19], such as image segmentation [20], face clustering [21-23], and motion segmentation [24, 25].

The core of the spectral clustering algorithm [26, 27] is building an affinity matrix, whose elements measure the similarity of corresponding data points. The pairwise distance-based method of building an affinity matrix measures the similarity of two data points by computing the distance between them, e.g., Euclidean distance, which could only get the local structure of datasets, being sensitive to noises and outliers. However, representation coefficients-based methods [28-32] are robust to noise and outliers, because the value of coefficient depends on both the two connected data points and the other data points. Recently, a few of works have shown that representation coefficient-based subspace clustering algorithms, such as robust subspace segmentation by low-rank representation (LRR) [18], sparse subspace clustering (SSC) [21], and constructing the L2-graph for subspace learning and subspace clustering (L2-graph) [32], perform better than those based on pairwise distance.

Corresponding author: Xiao-Jun Wu (wu_xiaojun@jiangnan.edu.cn)

Although LRR and SSC algorithms have achieved desirable clustering quality, they obtain the approximation solution of the coefficient matrix by iteration, and there is no more appropriate method to construct the affinity matrix, so there is a great room for improvement on running time and clustering accuracy of the algorithms. Affinity matrix is the core of the spectral clustering algorithm, whose elements measure a similarity degree of corresponding data points. Constructing a suitable affinity matrix can effectively improve the accuracy of algorithms. So, L2-graph subspace clustering algorithm takes the absolute value of each element of the coefficient matrix, retain the k largest coefficients per column, and set the other elements to 0, to get a new coefficient matrix, and then use the new coefficient matrix to construct a sparse affinity matrix. Moreover, the L2-graph algorithm can obtain a closed-form coefficient matrix, but it needs to encode each sample to obtain the coefficient matrix separately. So there is still much room for improvement. The FSSC clustering algorithm constructs a new objective function with a symmetric constraint based on F-norm, which can directly obtain the closed-form solution of the coefficient matrix by matrix calculation. Moreover, FSSC adds a symmetry constraint on the coefficient matrix, which can more really reflect the similarity between samples. The experimental results on face clustering and motion segmentation show FSSC algorithm can not only obviously reduce the running time, but also achieve higher accuracy.

The rest of the article is organized as follows: Section 2 presents the related work about representation-based subspace clustering algorithms, and Section 3 proposes the FSSC algorithm by constructing a new objective function with a symmetric constraint based on F-norm and provides the detailed derivation of obtaining the closed-form solution. Section 4 reports the results of a series of experiments to examine the effectiveness of the algorithm in the context of face clustering and motion segmentation. Finally, Section 5 concludes this work.

## 2. RELATED WORK

Exploring the structure of data space is a challenging task in a diverse set of fields, which often relates to a rank-minimization problem. Low-Rank Subspace Clustering (LRSC) ($P_1$) [22] solves the following optimization problem to get the coefficient matrix:

$$\min \|C\|_* + \tfrac{\tau}{2} \|Y\text{-}YC\|_F^2 \ \ \text{s.t.} \ \ C = C^T. \tag{1}$$

where $\|\cdot\|_*$ denotes the nuclear norm, i.e., the sum of the singular values of a matrix. $C \in \mathbb{R}^{n \times n}$ is the low-rank representation of the data set $Y \in \mathbb{R}^{m \times n}$. The procedure of the low-rank subspace clustering algorithm ($P_1$) is described in Algorithm 1.

| Algorithm 1   LRSC ($P_1$) [22] |
|---|

Input：a set of points $Y \in \mathbb{R}^{m \times n}$, and the number of clusters u.
1：Solve the nuclear norm minimization problem (1) to get $C = [c_1, c_2, \ldots, c_n]$.
2：Form an affinity matrix $W = |C| + |C|^T$.
3：Apply spectral clustering to the affinity matrix W.
Output: The cluster assignments of $Y$: $Y_1, Y_2, \ldots, Y_u$.

It shows that the collaborative representation-based subspace clustering algorithm can achieve better clustering quality than that got by the low-rank representation-based subspace clustering algorithm [32]. The L2-graph algorithm needs to solve the following problem,

$$\min_{c_i} \tfrac{1}{2} \|y_i - Y_i c_i\|_2^2 + \tau \|c_i\|_2^2. \tag{2}$$

where $c_i$ is the collaborative representation of the dataset $Y_i = [y_1, \ldots, y_{i-1}, 0, y_{i+1}, \ldots, y_n]$, The closed-form solution of (2) can be obtained easily by the Lagrange multiplier method. The implementation process of the L2-graph algorithm is described in algorithm 2.

Corresponding author: Xiao-Jun Wu (wu_xiaojun@jiangnan.edu.cn)

| Algorithm 2    L2-graph [32] |
|---|
| Input: a set of points $Y \in \mathbb{R}^{m \times n}$, the number of clusters u and the number of reserved coefficients k per column |
| 1: Solve the L2 norm minimization problem (2) to get $c_i$ and normalize $c_i$ to give a unit L2 norm $\hat{c}_i$. |
| 2: Form an coefficient matrix $C^* = [\hat{c}_1, \hat{c}_2, \ldots, \hat{c}_n]$. |
| 3: Take the absolute value of each element of the coefficient matrix, and then retain the k largest coefficients per column, set the other elements to 0, to get a new coefficient matrix $\hat{C}$. |
| 4: Form an affinity matrix $W = \left|\hat{C}\right| + \left|\hat{C}\right|^T$. |
| 5: Apply spectral clustering to the affinity matrix W. |
| Output: The cluster assignments of $Y$: $Y_1, Y_2, \ldots, Y_u$. |

## 3. Constructing the F-Graph with a Symmetric Constraint for Subspace Clustering

Before describing the FSSC algorithm in detail, a lemma [22] is introduced as follows:

Lemma1. For any real-valued, symmetric positive definite matrices $X \in \mathbb{R}^{n \times n}$ and $Z \in \mathbb{R}^{n \times n}$,

$$\text{trace}(XZ) \geqslant \sum_{i=1}^{n} \sigma_i(X)\sigma_{n-i+1}(Z)$$

where $\sigma_1(X) \geqslant \sigma_2(X) \geqslant \cdots \geqslant 0$ and $\sigma_1(Z) \geqslant \sigma_2(Z) \geqslant \cdots \geqslant 0$ are the descending singular values of $X$ and $Z$, respectively. The case of equality occurs if and only if it is possible to find a unitary matrix $U_X$ that simultaneously singular value-decomposes X and Z in the sense that

$$X = U_X \Sigma_X U_X^T$$
$$Z = U_X \Pi \Sigma_Z \Pi^T U_X^T$$

where $\Sigma_X$ and $\Sigma_Z$ denote the n*n diagonal matrices with the singular values of $X$ and $Z$, respectively, down in the diagonal in descending order, and $\Pi$ is a permutation matrix such that $\Pi \Sigma_Z \Pi^T$ contains the singular values of $Z$ in the diagonal in ascending order.

On the basis of further studying the LRSC and L2-graph algorithms in section 1, we construct a new objective function as follows

$$\min_{C} \tfrac{\tau}{2} \|Y - YC\|_F^2 + \tfrac{1}{2} \|C\|_F^2 \quad \text{s.t.} \quad C = C^T \tag{3}$$

where $C \in \mathbb{R}^{n \times n}$ is the representation matrix of the dataset $Y \in \mathbb{R}^{m \times n}$. The detailed derivation of obtaining the closed form solution referring to [32] is described as follows

Let $Y = U \Lambda V^T$ be the SVD of Y and $C = U_C \Delta U_C^T$ be the eigenvalue decomposition (EVD) of $C$, which can guarantee that $C$ is symmetric. The cost function of (3) reduces to

$$\tfrac{\tau}{2} \left\| U\Lambda V^T - U\Lambda V^T U_C \Delta U_C^T \right\|_F^2 + \tfrac{1}{2} \left\| U_C \Delta U_C^T \right\|_F^2$$

$$= \tfrac{\tau}{2} \left\| U\Lambda V^T (I - U_C \Delta U_C^T) \right\|_F^2 + \tfrac{1}{2} \left\| U_C \Delta U_C^T \right\|_F^2$$

$$= \tfrac{\tau}{2} \left\| U\Lambda V^T U_C (I - \Delta) U_C^T \right\|_F^2 + \tfrac{1}{2} \left\| U_C \Delta U_C^T \right\|_F^2$$

$$= \tfrac{\tau}{2} \left\| U(\Lambda V^T U_C (I - \Delta)) U_C^T \right\|_F^2 + \tfrac{1}{2} \left\| U_C \Delta U_C^T \right\|_F^2$$

$$= \tfrac{\tau}{2} \left\| \Lambda V^T U_C (I - \Delta) \right\|_F^2 + \tfrac{1}{2} \|\Delta\|_F^2$$

$$= \tfrac{\tau}{2} \|\Lambda W (I - \Delta)\|_F^2 + \tfrac{1}{2} \|\Delta\|_F^2$$

$$= \tfrac{\tau}{2} \text{trace}((\Lambda W (I - \Delta))^T \Lambda W (I - \Delta)) + \tfrac{1}{2} \cdot \text{trace}(\Delta^2)$$

Corresponding author: Xiao-Jun Wu (wu_xiaojun@jiangnan.edu.cn)

$= \frac{\tau}{2}\operatorname{trace}((\boldsymbol{I}-\Delta)^2\boldsymbol{W}^T\Lambda^2\boldsymbol{W}) + \frac{1}{2}\cdot\operatorname{trace}(\Delta^2)$

where $\boldsymbol{W} = \boldsymbol{V}^T\boldsymbol{U}_C$. In order to minimize this cost function, we need to first take the first item of the cost function into consideration, i.e.,

$\operatorname{trace}((\boldsymbol{I}-\Delta)^2\boldsymbol{W}^T\Lambda^2\boldsymbol{W})$,

Applying Lemma 1 to $\boldsymbol{X} = (\boldsymbol{I}-\Delta)^2 = \boldsymbol{U}_X(\boldsymbol{I}-\Delta)^2\boldsymbol{U}_X^T$ and $\boldsymbol{Z} = \boldsymbol{W}^T\Lambda^2\boldsymbol{W} = \boldsymbol{U}_X\Pi\Lambda^2\Pi^T\boldsymbol{U}_X^T$, we can obtain $\boldsymbol{U}_X = \boldsymbol{I}$ and $\boldsymbol{W}^T = \boldsymbol{U}_X\Pi = \Pi$, then there is a new cost function as follows

$\frac{\tau}{2}\sum_{i=1}^n \sigma_i((\boldsymbol{I}-\Delta)^2)\sigma_{n-i+1}(\boldsymbol{W}^T\Lambda^2\boldsymbol{W}) + \frac{1}{2}\cdot\sum_{i=1}^n \sigma_i(\Delta^2)$

$= \frac{\tau}{2}\sum_{i=1}^n \sigma_i((\boldsymbol{I}-\Delta)^2)\sigma_{n-i+1}(\Lambda^2) + \frac{1}{2}\cdot\sum_{i=1}^n \sigma_i(\Delta^2)$

Let the ith largest element in the diagonal of $(\boldsymbol{I}-\Delta)^2$ and $\Lambda^2$ be $(1-\delta_i)^2 = \sigma_i((\boldsymbol{I}-\Delta)^2)$ and $\nu_{n-i+1}^2 = \lambda_i^2 = \sigma_i(\Lambda^2)$, respectively. Then

$\min \frac{\tau}{2}\sum_{i=1}^n \sigma_i((\boldsymbol{I}-\Delta)^2)\sigma_{n-i+1}(\Lambda^2) + \frac{1}{2}\cdot\sum_{i=1}^n \sigma_i(\Delta^2)$

$= \min \frac{\tau}{2}\sum_{i=1}^n (1-\delta_i)^2\nu_i^2 + \frac{1}{2}\cdot\sum_{i=1}^n \delta_i^2$

we can independently solve for each $\delta_i$ to find the optimal $\Delta$ as

$\delta_i = \underset{\delta}{\operatorname{argmin}}\ \frac{\tau}{2}(1-\delta)^2\nu_i^2 + \frac{1}{2}\cdot\delta^2$

The closed form solution to this problem can be obtained as

$\delta_i = P_0(\lambda_{n-i+1}) = P_0(\nu_i) = \begin{cases} \frac{1}{1+\frac{1}{\tau\nu_i^2}} & \nu_i > 0 \\ 0 & \nu_i \leqslant 0 \end{cases}$

which can be written in matrix form as $\Delta = \Pi P_0(\Lambda)\Pi^T$. Therefore,

$\Pi^T\Delta\Pi = P_0(\Lambda) = \begin{bmatrix} \frac{1}{1+\frac{1}{\tau\Lambda_1^2}} & 0 \\ 0 & 0 \end{bmatrix}$

where $\Lambda = \operatorname{diag}(\Lambda_1, \Lambda_2)$ is partitioned according to the two sets $I_1 = \{i : \lambda_i > 0\}$ and $I_2 = \{i : \lambda_i \leqslant 0\}$.

Because $\boldsymbol{W} = \boldsymbol{V}^T\boldsymbol{U}_C$ and $\boldsymbol{W}^T = \Pi$, the optimal C is equivalent to

$\boldsymbol{C} = \boldsymbol{U}_C\Delta\boldsymbol{U}_C^T = \boldsymbol{V}\boldsymbol{W}\Delta\boldsymbol{W}^T\boldsymbol{V}^T = \boldsymbol{V}\Pi^T\Delta\Pi\boldsymbol{V}^T = \begin{bmatrix}\boldsymbol{V}_1 & \boldsymbol{V}_2\end{bmatrix} \begin{bmatrix} \frac{1}{1+\frac{1}{\tau\Lambda_1^2}} & 0 \\ 0 & 0 \end{bmatrix} \begin{bmatrix}\boldsymbol{V}_1 & \boldsymbol{V}_2\end{bmatrix}^T$

The symmetric constraint criterion can preserve the subspace structures of high-dimensional data and guarantee weight consistency for each pair of data points so that highly correlated data points of subspaces are represented together [28]. However, the closed-form solution $\boldsymbol{C}$ is not sparse and contains a large number of redundancy relations, which will reduce the accuracy of the algorithm. Therefore, after obtaining the coefficient matrix of the data set, referring to [32], we take the absolute value of each element of the coefficient matrix and retain the k largest coefficients per column, set the other elements to 0, to get a new coefficient matrix $\hat{\boldsymbol{C}}$. Then, FSSC performs spectral clustering over the new coefficient matrix as described in Algorithm 3.

| Algorithm 3  FSSC |
|---|
| Input: A set of points $\boldsymbol{Y} \in \mathbb{R}^{m \times n}$, the number of clusters u and the number of reserved coefficients k per column |
| 1: Solve the F norm minimization problem (3) to get $\boldsymbol{C}^* = [\boldsymbol{c}_1, \boldsymbol{c}_2, \ldots, \boldsymbol{c}_n]$. |
| 2: Take the absolute value of each element of the coefficient matrix, and then retain the k largest coefficients per column, set the other elements to 0, to get a new coefficient matrix $\hat{\boldsymbol{C}}$. |

Corresponding author: Xiao-Jun Wu (wu_xiaojun@jiangnan.edu.cn)

3：Form an affinity matrix $W = \left|\hat{C}\right| + \left|\hat{C}\right|^T$.

4：Apply spectral clustering to the affinity matrix $W$.

Output: The cluster assignments of $Y$: $Y_1, Y_2, \ldots, Y_u$.

## 4. Experiments

In this section, we use the subspace clustering Accuracy, Normalized Mutual Information (NMI) and Running Time to evaluate the performance of the FSSC algorithm in dealing with two computer vision tasks: face clustering and motion segmentation. We choose the state-of-the-art subspace clustering algorithms as a baseline, such as robust subspace segmentation by low-rank representation (LRR) [18], sparse subspace clustering (SSC) [21], and constructing the L2-graph for subspace learning and subspace clustering (L2-graph) [32]. All the experiments are implemented in Matlab R2013a and ran on a personal computer with Intel Core i3-3240 CPU and 8GB memory.

Datasets: We evaluate the performance of the algorithms for face clustering using three accessible image datasets, i.e., AR [33], Extended Yale B (ExYaleB) [34], and Multiple PIE (MPIE) [35]. The overview of these databases is provided in TABLE I. ExYaleB dataset contains the frontal face images of 38 individuals, where there are about 64 images for each subject. Moreover, the AR dataset consists of over 4000 face images of 126 individuals (70 male and 56 female), where 26 images consisting of 14 clean images, 6 images with sunglasses, and 6 images with scarves for each subject acquired under various expression, varying lighting conditions. As in [36], we randomly select a subset containing 1400 clean face images from 50 male and 50 female subjects. MPIE dataset contains the facial images of 337 subjects captured under 15 viewpoints and 19 illumination conditions in up to four recording sessions.

TABLE I
DATA SETS USED IN THE EXPERIMENTS

| Database | Samples | Original size | Cropped size | Features Dim. | classes |
|----------|---------|---------------|--------------|---------------|---------|
| AR | 1400 | 192*168 | 55*40 | 167 | 100 |
| ExYaleB | 2414 | 165*120 | 55*40 | 114 | 38 |
| MPIE-S2 | 2030 | 100*82 | 50*41 | 115 | 203 |
| MPIE-S3 | 1640 | 100*82 | 50*41 | 115 | 164 |
| MPIE-S4 | 1760 | 100*82 | 50*41 | 115 | 176 |

### 4.1 The influence of parameters on the FSSC algorithm

FSSC algorithm contains two parameters: the balance parameter τ and the number k of reserved coefficients per column, on which this section discusses the influence of the two parameters in AR and Extended Yale B datasets in detail.

Fig. 1 shows the evaluation results of FSSC with different values of the two parameters. In the AR dataset, when the τ of the FSSC algorithm ranges from 0.01 to 10, its accuracy varies from 37.21% to 85.14% and NMI varies from 66.31% to 93.40%. When the τ of FSSC ranges from 10 to 70, its Accuracy and NMI almost remain stable. When the k of FSSC algorithm ranges from 3 to 8, its Accuracy varies from 62.64% to 87.57% and NMI varies from 80.77% to 94.55%. With the increasing of k, its Accuracy and NMI has a tendency to decline. On the other hand, in the Extended Yale B dataset, we can find the same tendency that the performance of FSSC is superb and stable with the increasing of τ as in the AR dataset, and FSSC achieves the best result when k equals 6. Based on the above experimental results，we use τ=45 and k=8 for AR dataset, τ=3 and k=6 for the Extended Yale B dataset.

Corresponding author: Xiao-Jun Wu (wu_xiaojun@jiangnan.edu.cn)

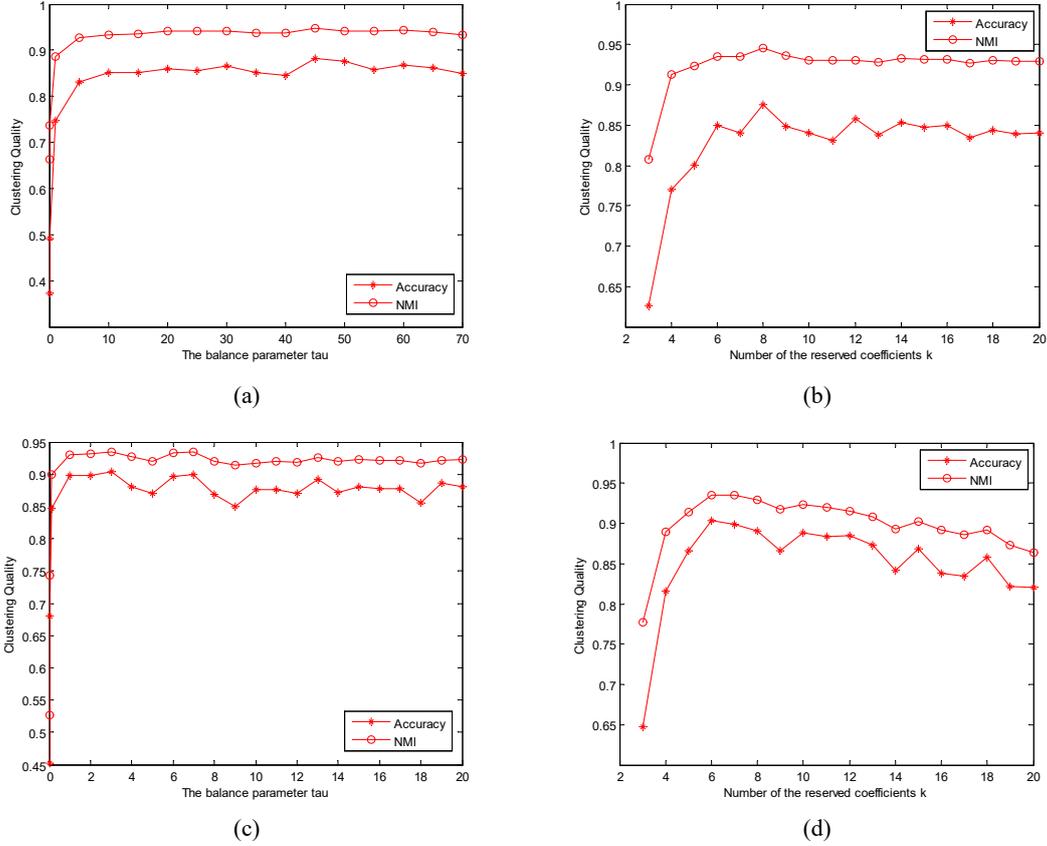

Fig. 1 The influence of parameters τ and k on the FSSC algorithm

Due to the limitation of space, we directly give parameter settings for MPIE and Hopkins 155 datasets. The parameters are set as τ=30 and k=13 for the MPIE dataset. In Hopkins 155 dataset, we use τ=26 and k=5 for 2 motions, τ=34 and k=5 for 3 motions.

**4.2 Face Clustering**

To evaluate the performance of the FSSC algorithm, we compare it with L2-graph, LRR, and SSC these latest subspace clustering algorithms in three aspects, including the clustering accuracy, NMI, and running time. For the state-of-the-art algorithms, we use the codes provided by their authors and set the parameters to be optimal. Results in Tables II and III are obtained by taking the mean after repeatedly running 20 times, and bold data represents the best performance.

TABLE II
CLUSTERING QUALITY(ACCURACY(AC(%)) AND THE CORRESPOIDING NORMALIZED MUTUAL INFORMATION(NMI(%))) OF DIFFERENT ALGORITHMS

| Databases | FSSC | | L2-graph [32] | | LRR [24] | | SSC [12] | |
|---|---|---|---|---|---|---|---|---|
| | AC | NMI | AC | NMI | AC | NMI | AC | NMI |
| AR | **86.94** | **94.31** | 83.43 | 92.56 | 75.36 | 89.80 | 77.56 | 88.40 |
| ExYaleB | **88.73** | **92.95** | 87.15 | 91.35 | 69.63 | 75.14 | 67.41 | 72.58 |
| MPIE-S2 | **90.00** | **97.92** | 89.17 | 97.47 | 84.05 | 96.58 | 86.66 | 97.25 |
| MPIE-S3 | **89.99** | **97.86** | 89.94 | 97.24 | 83.25 | 96.08 | 87.68 | 97.06 |
| MPIE-S4 | **89.68** | **97.68** | 89.39 | 97.66 | 83.97 | 96.44 | 86.37 | 97.06 |

Table II shows the clustering results of various approaches using different datasets. Our proposed FSSC algorithm has a distinct advantage in clustering quality compared with the other three subspace clustering algorithms. Construct the new objective function with a closed-form solution, and take the absolute value of each element of the coefficient matrix, and retain the k largest coefficients per column,

Corresponding author: Xiao-Jun Wu (wu_xiaojun@jiangnan.edu.cn)

set the other elements to 0, to get a new coefficient matrix, and then use the new coefficient matrix to construct a sparse affinity matrix is the key to the FSSC algorithm to achieve excellent performance. The four subspace clustering algorithms all perform good clustering quality on the MPIE dataset, where FSSC and L2-graph have almost the same clustering accuracy, whose accuracy is about 6% higher than that of LRR algorithm. On AR and ExYaleB datasets, the accuracy of the FSSC algorithm is about 3.51% and 1.58% higher than that of the L2-graph subspace clustering algorithm, respectively. The accuracy of LRR and SSC clustering algorithms on the two datasets is similar, but it is at least 10% lower than the accuracy of the FSSC algorithm.

TABLE III
AVERAGE RUNNING TIME (s) OF DIFFERENT ALGORITHMS

| Databases | FSSC | L2-graph [32] | LRR [24] | SSC [12] |
|---|---|---|---|---|
| AR | 70 | 78 | 73 | 138 |
| ExYaleB | 45 | 79 | 63 | 283 |
| MPIE-S2 | 213 | 231 | 233 | 375 |
| MPIE-S3 | 135 | 152 | 142 | 239 |
| MPIE-S4 | 135 | 158 | 153 | 271 |

TABLE III reports the time costs obtained by averaging the elapsed CPU time over 10 independent experiments for each algorithm. We can see that the FSSC algorithm achieves not only the optimal clustering quality but also the shortest running time. The running time of the FSSC algorithm is less about 9.7%, 42.85%, and 11.18% than that of L2-graph on AR, ExYaleB, and MPIE databases, respectively. Moreover, the L2-graph and LRR algorithms almost use the same running time. The reason is that FSSC can directly obtain the closed-form solution of the coefficient matrix by matrix calculation. L2-graph algorithm also can obtain a closed-form coefficient matrix. Still, it needs to encode each sample to get the coefficient matrix separately, and the LRR algorithm obtains the approximation solution of the coefficient matrix by iteration. Furthermore, SSC also achieves the approximation solution of the coefficient matrix by iteration, which costs about twice the running time than the FSSC algorithm.

**4.3 Motion segmentation**

Motion segmentation [22] refers to the problem of clustering a set of 2D feature points extracted from a video sequence into groups corresponding to different rigid-body motions. Here, the dimension of data matrix $Y$ is $2F*N$, where $N$ is the number of 2D feature trajectories, and $F$ is the number of frames in the video.

The Hopkins 155 dataset is used to evaluate the performance of the FSSC algorithm against that of the other algorithms, which consists of 120 video sequences of two motions and 35 sequences of three motions. Because the four subspace clustering algorithms all perform well in Hopkins 115 dataset, this section employs the subspace clustering error to evaluate the performance of the algorithms intuitively.

TABLE IV
CLUSTERING ERROR(%) OF THE EVALUATED ALGORITHMS ON THE HOPKINS 155 RAW DATA

| Algorithms | 2 Motions | | 3 Motions | | All | | |
|---|---|---|---|---|---|---|---|
| | Mean | Median | Mean | Median | Mean | Median | Run Time |
| FSSC | 1.54 | **0.00** | **2.61** | **0.00** | 1.78 | **0.00** | **32s** |
| L2-graph [32] | 2.30 | **0.00** | 4.78 | 0.47 | 2.86 | **0.00** | 42s |
| LRR [24] | 4.37 | 0.27 | 7.41 | 2.56 | 5.06 | 0.52 | 200s |
| SSC [12] | **1.53** | **0.00** | 4.40 | 0.56 | 2.18 | **0.00** | 212s |

Table IV shows the results of applying different subspace clustering algorithms to the original 2F

Corresponding author: Xiao-Jun Wu (wu_xiaojun@jiangnan.edu.cn)

dimensional feature trajectories. By analyzing the results, we can observe that the FSSC clustering algorithm is the best in clustering quality and running time, especially in the aspect of saving running time, less 23.8%, 84.9%, and 84% than the L2-graph, LRR, and SSC algorithm, respectively.

## 5. CONCLUSION

The structure of the objective function and affinity matrix is the core of the representation-based subspace clustering algorithm. On the basis of further studying the LRSC and L2-graph algorithms, we design an F-graph with a symmetric constraint algorithm for subspace clustering, which has a closed-form solution that can not only obviously reduce the running time, but also achieve higher accuracy. However, there are still many aspects worth further studying, such as the selection of the number k of reserved coefficients per column in the coefficient matrix, and the relationship between k and the intrinsic dimensionality of a subspace.

Corresponding author: Xiao-Jun Wu (wu_xiaojun@jiangnan.edu.cn)

Corresponding author: Xiao-Jun Wu (wu_xiaojun@jiangnan.edu.cn)